\documentclass[letterpaper, 10 pt, conference]{ieeeconf}  

\IEEEoverridecommandlockouts                              

\newif\ifenableanonymous
\enableanonymousfalse  

\usepackage[english]{babel}

\usepackage{cite}
\usepackage{graphicx}
\usepackage{subcaption}
\usepackage{multirow}
\usepackage{booktabs}
\usepackage{amsmath,amssymb}  
\usepackage[colorlinks=true, allcolors=blue]{hyperref}
\usepackage{utils/symbols}
\usepackage{algorithm}
\usepackage{algorithmicx}
\usepackage{algpseudocode}

\title{\LARGE \bf
PegasusFlow: Parallel Rolling-Denoising Score Sampling
\\ for Robot Diffusion Planner Flow Matching
}

\ifenableanonymous
\author{
    Anonymous Authors
}
\else
\author{
    Lei Ye, Haibo Gao, Peng Xu$^{*}$, Zhelin Zhang, Junqi Shan, Ao Zhang, Wei Zhang,\\Ruyi Zhou, Zongquan Deng and Liang Ding
    \thanks{The authors are with the State Key Laboratory of Robotics and Systems, Harbin Institute of Technology, Harbin 150001, China.}
    \thanks{*Corresponding author: Peng Xu (pengxu\_hit@163.com).}
    \thanks{This work was supported in part by the National Natural Science Foundation of China under Grant 52595712, Grant 52595710, Grant 52425502;
    the National Key R\&D Program of China (Grant No. 2024YFB4710900);
    the New Cornerstone Science Foundation through the XPLORER PRIZE.}
}
\fi

\begin{document}

\maketitle
\thispagestyle{empty}
\pagestyle{empty}

\begin{abstract}
    Diffusion models offer powerful generative capabilities for robot trajectory planning, yet their practical deployment on robots is hindered by a critical bottleneck: reliance on imitation learning from expert demonstrations. This paradigm is problematic as it is often impractical to produce high quality data for specialized robots, and it creates an inefficient, theoretically suboptimal training pipeline. To overcome this, we introduce PegasusFlow, a parallel rolling-denoising framework that enables direct  sampling of trajectory score gradients from environmental interaction, completely bypassing the need for expert data. Our core innovation is a sampling algorithm called Weighted Basis Function Optimization (WBFO), which leverages spline basis representations to achieve superior sample efficiency and faster convergence compared to traditional methods like MPPI. The framework is embedded within a scalable, asynchronous parallel simulation architecture that supports massively parallel rollouts for efficient data collection. Extensive experiments on trajectory optimization and robotic navigation tasks demonstrate that our approach, particularly Action-Value WBFO (AVWBFO) combined with a reinforcement learning warm-start, significantly outperforms baselines. In a challenging barrier-crossing task, our method achieved a 100\% success rate and was 18\% faster than the next-best method, validating its effectiveness for complex terrain locomotion planning.
    \ifenableanonymous
        \url{https://anonymous.4open.science/w/pegasusflow_page/}
    \else
        \url{https://masteryip.github.io/pegasusflow.github.io/}
    \fi
\end{abstract}


\section{Introduction}

Diffusion models have emerged as a powerful tool for trajectory generation in robotics, demonstrating a remarkable capability for refining motion plans in high-dimensional task spaces and handling complex constraints~\cite{2022Tevet_MDM, 2023Chi_DiffusionPolicy,2024Huang_DiffuseLoco, 2025Liao_BeyondMimicMotionTracking,2024Tevet_CLoSD,2024Disney_RobotMDM}.
As for legged robot locomotion, this translates into a potential to solve intricate navigation tasks that challenge traditional reinforcement learning (RL) policies, such as maneuvering in confined spaces with obstacles~\cite{2025Xu_PARCPhysicsbasedAugmentation}. However, the practical application of these diffusion planners is severely hampered by their reliance on extensive, high-quality expert demonstration data.

The predominant training paradigm for diffusion planners is imitation learning, where policies are trained via behavior cloning on datasets of expert trajectories, often collected from motion capture or generated by a pre-trained RL policy. This data-driven approach is problematic. 
First, collecting high-quality expert demonstrations is often effort-intensive, costly, and sometimes infeasible, particularly for specialized robots like hexapods where such data is scarce.
Second, this indirect training process, generating data with one policy (e.g., RL) to train another (the diffusion model), is theoretically suboptimal, computationally wasteful, and can introduce distribution shifts that degrade performance. This naturally motivates our idea: instead of learning from data, we need methods to estimate trajectory score gradients directly from environmental interactions, enabling more direct and efficient score-matching.

\begin{figure}[t]
    \centering
    \includegraphics[width=1\linewidth]{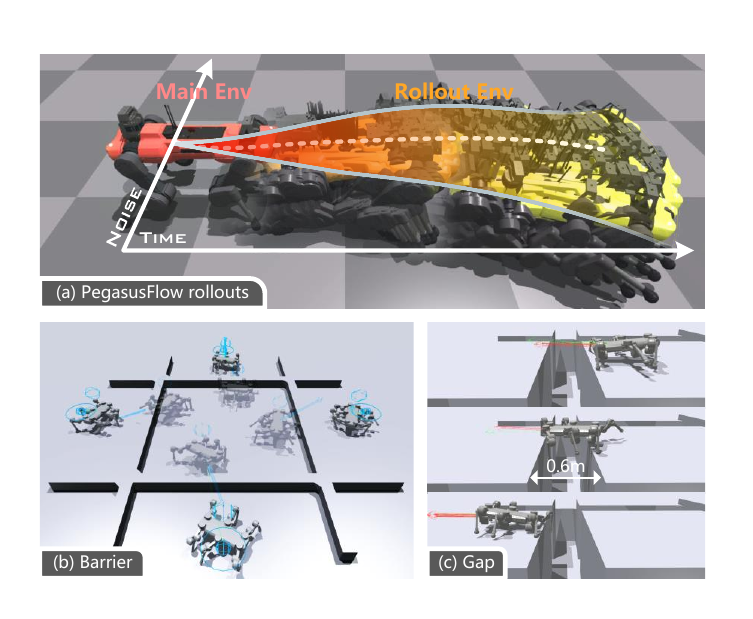}
    \caption{Illustration of PegasusFlow. (a) Visualization of the rollouts of an environments within a predict horizon. (b, c) The algorithm can optimize the robot control trajectory to navigate complex environments with barriers and gaps.}
    \label{fig:PegasusFlowDemos}
\end{figure}

Existing methods that touch upon direct score estimation, such as those inspired by Model Predictive Path Integral (MPPI) control~\cite{2024Xue_DialMPC,2024Pan_MBD}, are a step in the right direction but fall short in efficiency. They often require thousands of simulation rollouts to produce a single high-quality trajectory, rendering them unsuitable for parallel data generation. Furthermore, to the best of our knowledge, there is currently no known framework capable of performing this score sampling process in a massively parallel manner, which is essential for efficient data collection and diffusion planner training.

To address these gaps, we introduce a framework(Fig.~\ref{fig:PegasusFlow}) for direct, parallel, and efficient sampling of trajectory score gradients. Our approach formulates trajectory planning as a discrete-time optimal control problem and leverages a new sampling algorithm, Weighted Basis Function Optimization (WBFO), which provides superior sample efficiency and faster convergence than existing methods. To enable parallel data collection, we embed this optimizer within a scalable, asynchronous parallel simulation architecture built on IsaacGym. This system supports massively parallel rollouts, warm-starting with RL trajectories, and leverages NVIDIA Warp for accelerated Signed Distance Field (SDF) and raycasting queries.

The primary contributions of this work are:
\begin{enumerate}
    \item \textbf{Parallel Score Sampling Framework:} We present a framework to directly sample robot trajectory score gradients in parallel, which makes pure score-matching training without expert demonstrations and bridging diffusion generative models with real-time robotic control more promising. This includes a scalable mini-batch environment simulation system and a rolling-denoising optimization process using IsaacGym with coordinated main and rollout-environment architecture.
    \item \textbf{Weighted Basis Function Optimization (WBFO):} We develop an efficient trajectory optimization approach through spline basis functions with node-level importance weighting, achieving computational efficiency while maintaining trajectory quality through local optimality and temporal coherence.
    \item \textbf{Efficient Noise Sampling Schema:} We introduce an MPC-inspired structured noise scheduling approach combining Latin Hypercube Sampling(LHS), Hierarchical Ramp Noise Scheduling(HRNS) and RL Policy Warm Start for improved space exploration and convergence in trajectory optimization tasks.
\end{enumerate}
\section{Related Works}

\subsection{Trajectory Optimization and Model Predictive Control}

Model predictive control (MPC) algorithms search for functional gradients and perform gradient descent to solve optimal control problems\cite{2018DiCarlo_CMPCCheetah,2021Sleiman_UnifiedMPCFramework,2022Grandia_PerceptiveNMPC,2023Xu_RobustConvexModel}, they implement receding horizon control with rolling updates for real-time adaptation. Direct methods like Direct Collocation and Multiple Shooting transcribe continuous control into discrete nonlinear programs. While Di Carlo et al. \cite{2018DiCarlo_CMPCCheetah} and Grandia et al. \cite{2022Grandia_PerceptiveNMPC} achieved impressive results, these methods suffer from local minima and may fail in complex environments. Contact planning methods in our previous work \cite{MCTS,KCFRC} address foothold selection for legged robots but operate at discrete contact level rather than continuous trajectory optimization.

As for sampling-based methods, MPPI\cite{2016Williams_MPPI} and its variants \cite{2022Yin_TrajectoryDistributionControl,2025Pezzato_SamplingbasedModelPredictive,2024Yi_CoVOMPCTheoreticalanalysis} perform trajectory sampling based on optimal control theory, using relationships between free energy and relative entropy. However, MPPI requires extensive computation during rollout, limiting its capabilities to extend to multiple robot environments.

These concepts inspire our rolling-denoising framework, leveraging MPC's receding horizon principle while addressing computational limitations through hierarchical noise scheduling and parallel batch rollout.

\subsection{Score-based Models for Robot Planning}

DDPM \cite{2020Ho_DDPM} establishes diffusion models through gradual noise reversal, while DDIM \cite{2022Song_DDIM} achieves 10-50x acceleration via non-Markovian paths. SDE formulation \cite{2021Song_SDE,2023Song_ConsistencyModels} provides continuous-time generalization with connections to stochastic calculus.

Diffusion Policy \cite{2023Chi_DiffusionPolicy} handles multimodal action distributions for manipulation tasks. Vision-Language-Action models like $\pi_0$ \cite{2024_PI0} and $\pi_{0.5}$ \cite{2025_Pi05} demonstrate broad generalization. These approaches excel in structured environments with moderate computational requirements.

For legged locomotion, DiffuseLoco \cite{2024Huang_DiffuseLoco,2025Liao_BeyondMimicMotionTracking} enables real-time control from offline datasets but relies on expert trajectories. Model-Based Diffusion \cite{2024Pan_MBD} and Dial-MPC \cite{2024Xue_DialMPC} eliminate expert data dependency but lack batch processing capabilities. Several works integrate diffusion motion generator and RL tracking policies for downstream tasks \cite{2024Disney_RobotMDM,2024Tevet_CLoSD,2025Xu_PARCPhysicsbasedAugmentation,huang2025diffusecloc}.

Diffusion policies are relatively less used in legged locomotion over challenging terrain due to: (1) real-time control requirements conflicting with low update frequency, (2) computational cost of iterative denoising, and (3) RL's effectiveness for basic locomotion. However, diffusion offers superior potential for complex environment navigation and obstacle avoidance, motivating our direct score gradient sampling approach.








\begin{figure*}[t!]
    \centering
    \includegraphics[width=1.0\textwidth]{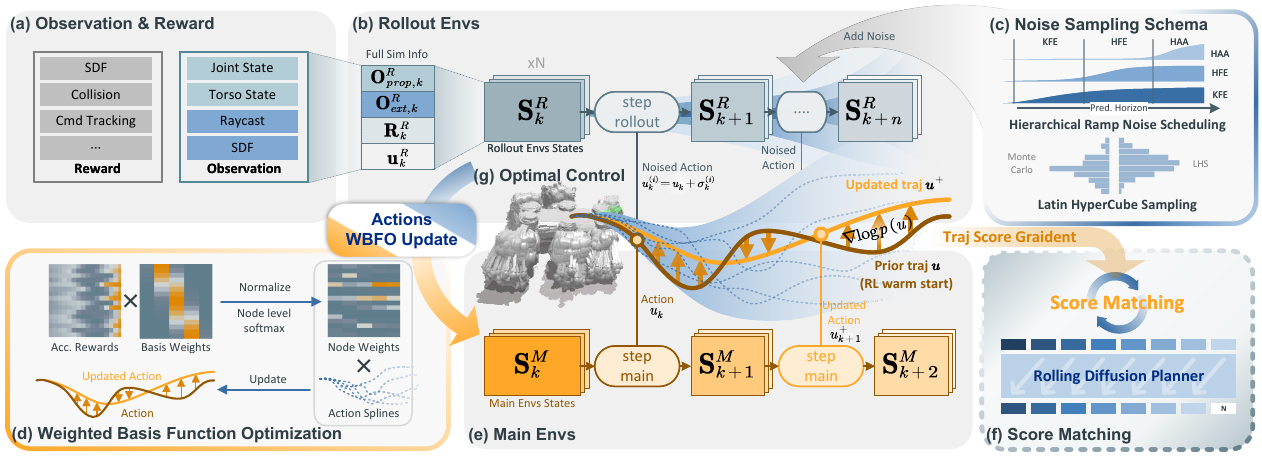}
    \caption{Framework of PegasusFlow. (a) RL style observations and rewards setup. (b) Rollout environments for trajectory score gradient sampling. (c) Noise sampling schema (Section \ref{subsection:noise_sampling_schema}). (d) WBFO for trajectory optimization (Section \ref{subsection:wbfo}) (e) Main environments that interact with simulation. (f) Sampled trajectory score gradient can be used for flow matching training of diffusion policy. (g) Optimal control problem and sampling-based score gradient estimation.}
    \label{fig:PegasusFlow}
\end{figure*}

\section{Preliminary}
\label{sec:problem_formulation}

\subsection{Problem Formulation}

We formulate the robot planning problem as a discrete-time optimal control problem, depicted in Fig.~\ref{fig:PegasusFlow}(g). The system must achieve desired objectives while satisfying constraints and maintaining operational requirements.

The discrete-time system dynamics are given by:
\begin{equation}
    x_{k+1} = f(x_k, u_k, t_k)
\end{equation}
where $x_k \in \mathbb{R}^n$ is the state vector at time step $k$ containing the robot's position, orientation, velocities, and joint configurations, $u_k \in \mathbb{R}^m$ is the control input vector, and $f(\cdot)$ represents the nonlinear system dynamics function.

The finite-horizon optimal control problem is formulated as:
\begin{equation}
    \begin{gathered}
        \min_{u_{i:i+h}} J(u_{i:i+h}) = e^{-\alpha (h+1) \Delta t} l_f(x_{i+h+1}) \\
        + \sum_{k=i}^{i+h} e^{-\alpha \Delta t (k-i)} \cdot l(x_k,u_k,t_k;c_k,o_k) \Delta t \\
        \text{s.t.} \quad x_{k+1} = f(x_k, u_k, t_k)
    \end{gathered}
\end{equation}
where $h$ is the planning horizon, $l_f(\cdot)$ is the terminal cost function, $\alpha > 0$ is the exponential discount factor that prioritizes immediate costs while maintaining long-term planning capability, $c_k$ represents command inputs at time $k$, and $o_k$ represents environmental observations.

The stage cost function $l(x_k,u_k,t_k;c_k,o_k)$ includes multiple components:
\begin{equation}
    \begin{gathered}
        l(x_k,u_k,t_k;c_k,o_k) = l_{\text{task}}(x_k,t_k;c_k) \\ + l_{\text{obstacle}}(x_k;o_k) + l_{\text{control}}(u_k)
    \end{gathered}
\end{equation}
where $l_{\text{task}}$ represents task-related costs such as trajectory tracking error with respect to commands $c_k$, $l_{\text{obstacle}}$ penalizes constraint violations based on observations $o_k$, and $l_{\text{control}}$ penalizes excessive control effort to ensure smooth operation.

\subsection{Estimating Score Gradient using MPPI}

For high-dimensional systems with complex dynamics, directly solving the optimal control problem is computationally intractable. Instead, MPPI can not only approximates the optimal control through stochastic sampling, but also estimate trajectory score gradient from sampled trajectories.

MPPI works by adding noise to the control inputs and evaluating the resulting trajectories, as is shown in Fig.~\ref{fig:PegasusFlow}(g). Specifically, we sample $N$ noisy control sequences:
\begin{equation}
    u_k^{(i)} = u_k + \sigma_k^{(i)}
\end{equation}
where $\sigma_k^{(i)}$ represents zero-mean Gaussian noise added to the nominal control $u_k$ for the $i$-th sample trajectory.

The MPPI approximation of the optimal control policy is then given by the importance-weighted average:
\begin{equation}
    u^+_{i:i+h} = \frac{\sum_{j=1}^N u^{(j)}_{i:i+h} e^{-\frac{1}{\lambda}J^{(j)}}}{\sum_{j=1}^N e^{-\frac{1}{\lambda}J^{(j)}}}
\end{equation}
where $\lambda > 0$ is a temperature parameter that controls the exploration-exploitation trade-off. A smaller $\lambda$ leads to more aggressive selection of low-cost trajectories, while a larger $\lambda$ provides more exploration.

An important theoretical insight from \cite{2024Xue_DialMPC} is that the MPPI update is mathematically equivalent to a single-stage diffusion process. The MPPI update can be viewed as a one-step gradient ascent using the score function $\nabla \log p(u)$:
\begin{equation}
    u^+ = u + \Sigma \nabla \log p_1(u)
\end{equation}
where $p_0(u) \propto \exp(-\frac{J(u)}{\lambda})$ represents the probability distribution induced by the cost function, and $p_1(\cdot) \propto (p_0 * \phi)(\cdot)$ is the smoothed distribution with kernel $\phi$ of the introduced gaussian noise $\sigma_k$.

\section{Sampling-based Optimization Method}
\label{sec:wbfo}

The connection between MPPI and diffusion processes provides the theoretical foundation for our approach (Fig.~\ref{fig:PegasusFlow}(f)). However, standard MPPI requires sampling a large number of trajectories $N$ to achieve good approximation quality, which can be computationally expensive for batch robot simulation. This limitation motivates our development of more efficient sampling strategies based on the WBFO algorithm described in the following sections(Fig.~\ref{fig:PegasusFlow}(c,d)).

\subsection{Spline Trajectory Representation}
\label{sec:spline_trajectory_representation}

In this paper, we employ cubic spline interpolation for trajectory representation, which provides smooth, continuous trajectories through control nodes while maintaining computational efficiency. Consider a local parameter $t \in [0,1]$ representing the normalized position within a trajectory segment.

The position at parameter $t$ can be computed using a general cubic spline formulation:
\begin{equation}
    \mathbf{Pos}(t) = \mathbf{P} \cdot \mathbf{M} \cdot \mathbf{T}
\end{equation}
where:
\begin{itemize}
    \item $\mathbf{P} = [P_{i-1}, P_i, P_{i+1}, P_{i+2}]$ is the matrix of control points influencing the current segment
    \item $\mathbf{M}$ is the cubic spline basis matrix that defines the interpolation scheme
    \item $\mathbf{T} = [1, t, t^2, t^3]^T$ is the power basis vector
\end{itemize}
For our implementation, we specifically choose Catmull-Rom splines due to their interpolation property, which ensures that the trajectory passes exactly through all control nodes.
The spline representation allows us to construct a basis weight matrix $\mathbf{\Phi}$ that maps directly from control nodes to trajectory samples. This matrix enables efficient conversion between the two representations. For $D$ dimension trajectories with $K$ control nodes and $T$ discrete time steps, the relationship can be expressed as:
\begin{equation}
    \boldsymbol{u} = \mathbf{P} \cdot \mathbf{\Phi}
\end{equation}
where $\boldsymbol{u} \in \mathbb{R}^{D \times T}$ represents the dense trajectory samples, $\mathbf{\Phi} \in \mathbb{R}^{K \times T}$ is the precomputed basis weight matrix, and $\mathbf{P} \in \mathbb{R}^{D \times K}$ contains the control nodes.

For the reverse operation, we compute the pseudo-inverse of the basis weight matrix $\mathbf{\Phi}^+$ to minimize the reconstruction error:
\begin{equation}
    \mathbf{P} = \boldsymbol{u} \cdot \mathbf{\Phi}^+
\end{equation}
This approach ensures that the conversion between discrete samples and control nodes is mathematically optimal, minimizing the mean squared error between the original dense trajectory and the reconstructed trajectory from the control nodes.

\subsection{Trajectory Sampling through Basis Function}

\begin{figure}
    \centering
    \includegraphics[width=1\linewidth]{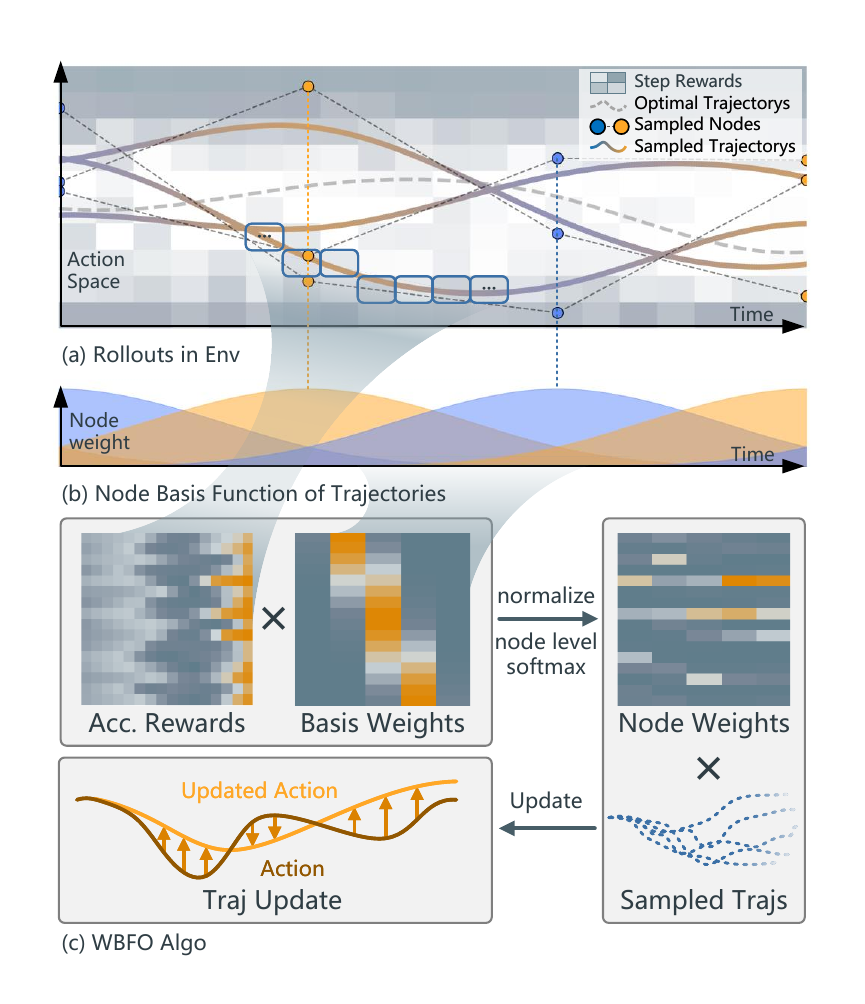}
    \caption{Weighted Basis Function Optimization. (a) Sampled action trajectories with step-wise rewards. (b) Basis functions of the bundle of the sampled trajectories. (c) Schematic diagram of the WBFO process, illustrating the mapping from trajectory rewards to node weights via basis functions.}
    \label{fig:wbfo_sch}
\end{figure}

Building on the spline representation, we can sample the functional space of the trajectory more efficiently by utilizing basis functions and step rewards. Instead of sampling complete trajectories directly, we sample the control points $\{P_k\}$ in the reduced-dimensional space, which results in more efficient exploration because:

\begin{enumerate}
    \item The dimensionality of the control point space ($K$ nodes) is typically much lower than the full trajectory space ($T$ timesteps), where $K \ll T$.
    \item Each control point influences only a local portion of the trajectory through its associated basis functions, enabling targeted optimization.
    \item The basis function representation inherently enforces smoothness constraints, eliminating physically implausible trajectories.
\end{enumerate}

Mathematically, spline trajectory $u(t)$ can be represented as a linear combination of basis functions:
\begin{equation}
    u(t) = \sum_{k=1}^{K} P_k \phi_k(t)
\end{equation}
where $\{P_k\}_{k=1}^{K}$ are the control point coefficients and $\{\phi_k(t)\}_{k=1}^{K}$ are the basis functions. The optimization problem becomes finding optimal control points:
\begin{equation}
    P_k^* = \arg\min_{P_k} \mathbb{E}_{u \sim p(u | \{P_k\})} \left[ \sum_{t=1}^{T} r_t \right]
\end{equation}
where $r_t$ is the reward at time $t$ and the expectation is taken over trajectories sampled based on the control points $\{P_k\}$ instead of sampling complete trajectories directly like MPPI.
This functional space sampling exploits the low-dimensional structure of the trajectory manifold for efficient optimization.

\subsection{Weighted Basis Function Optimization}
\label{subsection:wbfo}

We propose Weighted Basis Function Optimization (WBFO), which extends conventional MPPI by decomposing trajectories into control points and applying spatiotemporal weighting through basis function representations. This approach enables targeted optimization of trajectory segments while maintaining global coherence.

As illustrated in Fig.~\ref{fig:wbfo_sch}, our unified WBFO framework(\textbf{Algorithm \ref{alg:unified_wbfo}}) handles both trajectory optimization ($\gamma = 0$) and optimal control ($\gamma > 0$) through a discount factor parameter. The algorithm computes discounted accumulated rewards from step-wise rewards collected from rollout environments in Fig.~\ref{fig:wbfo_sch}(a):
\begin{equation}
    \label{eq:discounted_reward}
    \mathbf{R}_{\text{acc}}(i,t) = \sum_{s=t}^{T} r_{i,s} \cdot \gamma^{s-t}
\end{equation}
where $r_{i,s}$ is the reward of trajectory $i$ at timestep $s$. When $\gamma = 0$, this reduces to step-wise rewards $\mathbf{R}_{\text{acc}}(i,t) = r_{i,t}$ for trajectory optimization. When $\gamma > 0$, it provides temporal credit assignment for systems with integrators.

The core innovation lies in computing node-specific weights through basis function mapping using basis function in Fig.~\ref{fig:wbfo_sch}(b):
\begin{equation}
    \label{eq:node_weight}
    \mathbf{W} = \mathbf{R}_{\text{acc}} \mathbf{\Phi}^T
\end{equation}
where $\mathbf{\Phi} \in \mathbb{R}^{K \times T}$ maps timesteps to control nodes, and $\mathbf{R}_{\text{acc}} \in \mathbb{R}^{N \times T}$ contains trajectory rewards. Node-level normalization ensures balanced updates:
\begin{equation}
    \label{eq:node_normalization}
    \mathbf{W}(i,j) = \frac{\mathbf{W}(i,j) - \text{mean}(\mathbf{W}(i,:))}{\text{std}(\mathbf{W}(i,:))}
\end{equation}

\begin{algorithm}[h]
    \caption{Unified Weighted Basis Function Optimization}
    \label{alg:unified_wbfo}
    \renewcommand{\algorithmicrequire}{\textbf{Input:}}
    \renewcommand{\algorithmicensure}{\textbf{Output:}}
    \begin{algorithmic}[1]
        \Require nodes $\mathbf{P} \in \mathbb{R}^{D \times K}$ of prior $\boldsymbol{u}$ , noise schedule $\sigma \in \mathbb{R}^{D \times K}$, discount factor $\gamma$, number of samples $N$
        \Ensure Updated control points $\mathbf{P}^+ \in \mathbb{R}^{D \times K}$
        \State Sample $N$ trajectories $\mathbf{U} \in \mathbb{R}^{N \times T \times D}$ from prior $\boldsymbol{u}$ by adding noise $\sigma$ to $\mathbf{P}$
        \State Evaluate step-wise rewards $\mathbf{R} \in \mathbb{R}^{N \times T}$ for all trajectories by interacting using rollout environments in Fig.~\ref{fig:PegasusFlow}(b)
        \State Calculate basis function mask $\boldsymbol{\Phi} \in \mathbb{R}^{K \times T}$
        \If{$\gamma > 0$} \Comment{AVWBFO-Integrated Systems}
        \State Compute accumulated rewards $\mathbf{R}_{\text{acc}}$ \Comment{Eq. (\ref{eq:discounted_reward})}
        \Else \Comment{WBFO}-Traj Optimization
        \State $\mathbf{R}_{\text{acc}} \leftarrow \mathbf{R}$ \Comment{Use step-wise rewards directly}
        \EndIf
        \State Calculate node weight matrix $\mathbf{W} = \mathbf{R}_{\text{acc}} \boldsymbol{\Phi}^T$ \Comment{Eq. (\ref{eq:node_weight})}
        \State Apply node-level normalization to $\mathbf{W}$ \Comment{Eq. (\ref{eq:node_normalization})}
        \State Compute node-level softmax weights: $w_{ij} = \frac{\exp(\mathbf{W}(i,j))}{\sum_{k} \exp(\mathbf{W}(i,k))}$
        \State Update control points: $P_j^+ = \sum_{i=1}^{N} w_{ij} \cdot P_{ij}$
        \State \Return $\mathbf{P}^+ = [P_1^+, P_2^+, \ldots, P_K^+]$
    \end{algorithmic}
\end{algorithm}

This unified approach offers several advantages over conventional MPPI:
\begin{itemize}
    \item \textbf{Local Optimality:} By applying node-level focus, we can concentrate optimization efforts on challenging segments of the trajectory.
    \item \textbf{Temporal Coherence:} The basis function mask enforces smoothness and temporal consistency across the optimized trajectory.
    \item \textbf{Efficient Exploration:} The weighted update scheme enables more effective exploration of the state space in regions of high uncertainty.
    \item \textbf{Unified Framework:} A single algorithm handles both trajectory optimization and optimal control problems through the discount factor parameter.
\end{itemize}

The WBFO approach forms a critical component of our rolling diffusion planner, enabling efficient trajectory refinement while maintaining the theoretical foundations of stochastic optimal control.

\subsection{Noise Sampling Schema}
\label{subsection:noise_sampling_schema}

We propose a dual-component approach combining Latin Hypercube Sampling (LHS) with Hierarchical Ramp Noise Scheduling (HRNS) to achieve superior sample efficiency and space exploration, shown in Fig.~\ref{fig:PegasusFlow}(c).

\textbf{Latin Hypercube Sampling.} Traditional Monte Carlo sampling can lead to clustering and poor coverage in high-dimensional trajectory optimization problems. LHS addresses this by ensuring uniform distribution across all noise dimensions through stratified sampling, where each dimension is divided into equally probable intervals with exactly one sample per interval. This guarantees better parameter space coverage with fewer samples, translating to more diverse exploration of motion primitives.

\textbf{Hierarchical Ramp Noise Scheduling.} HRNS introduces a structured approach to noise level management in receding horizon planning. The scheduling follows a ramp structure where noise magnitudes are modulated according to temporal distance from the current planning step, enabling focused exploration in critical near-term decisions while maintaining broader exploration for long-term planning.

The combination creates a synergistic effect: LHS ensures comprehensive space exploration while HRNS directs this exploration toward critical regions of the planning horizon, resulting in faster convergence and improved solution quality.



\section{Asynchronous Parallel Simulation}

We design an asynchronous parallel simulation framework for diffusion-based trajectory optimization that enables efficient data collection and score sampling through specialized adaptations for sampling from hundreds of sub-environments, shown in Fig.~\ref{fig:PegasusFlow}.

\subsection{Batch Rollouts in Parallel}

Our parallel rollout system implements a hierarchical environment structure for efficient trajectory sampling, which is illustrated in Fig.~\ref{fig:PegasusFlow}(b,e). The framework maintains:

\begin{equation}
    N_{\text{total}} = N_m \times (1 + N_r)
\end{equation}
where main environments $N_m$ are positioned at regular intervals among rollout environments $N_r$: $\{0, 1+N_r, 2(1+N_r), \ldots\}$.

The system executes two key operations: \textbf{state synchronization} resets rollout environments to match main environment states before each rollout; \textbf{rollout execution} allows parallel exploration of trajectory branches from identical initial states while main environments remain frozen. After rollout completion, main environments are restored from their cached states to continue execution, ensuring temporal consistency for receding horizon planning.

\subsection{Rolling-Denoising Optimization}

Our framework employs a rolling-denoising optimization strategy that integrates trajectory optimization within the receding horizon control loop (Fig.~\ref{fig:PegasusFlow}(b,e)). At each optimization interval, the system performs $n_{\text{denoise}}$ denoising steps (i.e., the number of denoising iterations) on the current trajectory nodes using the WBFO algorithm described in Section \ref{subsection:wbfo}.

The optimization process follows a Model Predictive Control (MPC) paradigm: (1) the current node trajectories are optimized through multiple denoising iterations using parallel rollout environments; (2) the optimized trajectory is executed for one timestep in the main environments; (3) trajectories are shifted forward in time, with new terminal actions appended via RL policy warm start in Section \ref{subsection:RLWarmStart}; (4) the cycle repeats with updated environmental observations.

\subsection{RL Policy Warm Start}
\label{subsection:RLWarmStart}

We implement an RL policy warm start mechanism that initializes optimization with high-quality action sequences from pre-trained policies.
The warm start process involves: \textbf{Trajectory initialization}. The program executes rollout procedures using the loaded policy to generate initial action sequences covering the planning horizon; \textbf{trajectory append} For receding horizon planning, the RL policy generates the last actions for appending during trajectory shifting operations. For convenience, we use RL style observations and rewards setup both for RL warm start and trajectory optimization as shown in Fig.~\ref{fig:PegasusFlow}(a).












\section{Experiments}

We validate the proposed hierarchical rolling-denoising framework through three key experimental aspects: (1) validating that PegasusFlow enables effective parallel trajectory score evaluation across diverse robotic tasks, (2) demonstrating that WBFO/AVWBFO (proposed) optimization algorithms combined with Latin Hypercube Sampling (LHS) achieve superior sample efficiency over MPPI baselines, and (3) showing that RL warm-starting significantly enhances trajectory optimization performance.
It should be noted that in all experiments, \textbf{$\gamma = 1.0$ is used for AVWBFO (proposed)}.
These experiments collectively demonstrate how our framework components(advanced optimization algorithms, strategic sampling, and RL integration) synergistically improve trajectory planning efficiency and success rates.

\subsection{Various Tasks in PegasusFlow}

\begin{figure}
    \centering
    \includegraphics[width=1\linewidth]{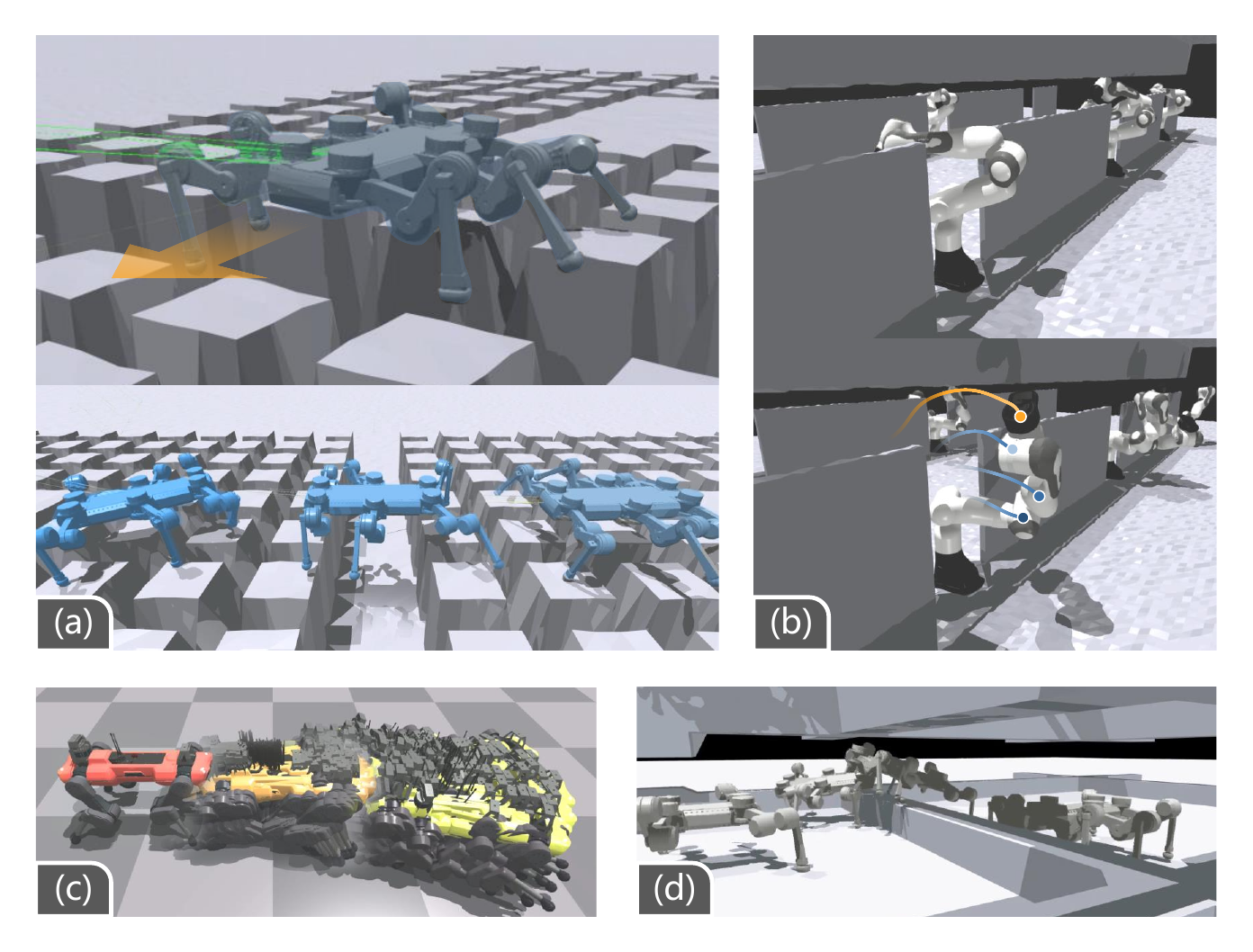}
    \caption{Robotic tasks in PegasusFlow. (a) Hexapod timber piles navigation. (b) Franka arm collision avoidance planning. (c) Quadruped walking. (d) Hexapod confined space navigation. Videos can be found in the supplementary material or at
    \ifenableanonymous
        \href{https://anonymous.4open.science/w/pegasusflow_page/}{this page}
    \else
        \href{https://masteryip.github.io/pegasusflow.github.io/}{this page}
    \fi.}
    \label{fig:exp_demos}
\end{figure}

We validate PegasusFlow's versatility across diverse robotic platforms and challenging scenarios, as illustrated in Fig.~\ref{fig:exp_demos}. Our framework demonstrates robust performance in several distinct domains: (a) \textbf{Hexapod timber piles navigation} tests complex terrain traversal with irregular obstacles requiring precise foothold planning and dynamic balance, (b) \textbf{Franka arm collision avoidance} validates high-dimensional manipulation planning in cluttered environments with safety constraints, (c) \textbf{Quadruped walking} shows hundreds of rollout environments for trajectory score sampling, and (d) \textbf{Hexapod confined space navigation} evaluates performance in spatially constrained environments where traditional planning methods typically fail.

\subsection{Optimization Algorithm Performance}

\subsubsection{Algorithm Comparison under Varying Sample Numbers}
We evaluate WBFO/AVWBFO (proposed) performance against MPPI baselines across two representative scenarios: 2D Navigation (point-to-point trajectory optimization with 25 random circular obstacles in a 10×10 meter workspace) and Inverted Pendulum (optimal control for system with integrators). Three optimization approaches are compared: (1) \textit{AVWBFO (proposed)}, (2) \textit{WBFO (proposed)}, and (3) \textit{MPPI} baseline.

For 2D Navigation, trajectories use 16 nodes interpolated to 64 dense samples over 10 optimization iterations with exponential noise decay (initial noise 3.0, decay rate 0.6). For Inverted Pendulum, trajectories use 32 nodes interpolated to 128 dense samples with identical optimization parameters. Statistical significance is ensured through 5 independent trials.

\begin{figure}
    \centering
    \includegraphics[width=1\linewidth]{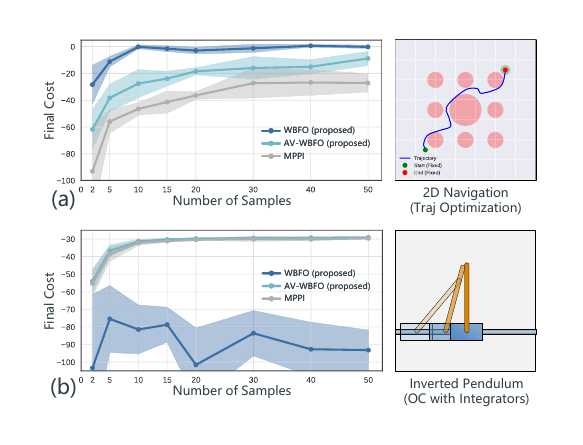}
    \caption{Optimization performance comparison. (a) 2D Navigation (trajectory planning problem). (b) Inverted Pendulum (optimal control problem). X-axis shows the number of samples used, Y-axis shows the final cost after optimization iterations.}
    \label{fig:exp1_samples_cost}
\end{figure}

\textbf{Results.} As shown in Fig.~\ref{fig:exp1_samples_cost}, our proposed methods demonstrate different performance characteristics across the two experimental scenarios:

In trajectory optimization tasks (2D Navigation, Fig.~\ref{fig:exp1_samples_cost}(a)), WBFO (proposed) demonstrates superior sample efficiency compared to MPPI, particularly with limited samples ($\le$20). For example, with 10 samples, WBFO (proposed) achieves a final cost of -0.095$\pm$1.46, while MPPI achieves -46.53$\pm$4.60, indicating WBFO's better performance with fewer samples. AVWBFO (proposed) performs intermediately between WBFO (proposed) and MPPI in trajectory optimization.

In optimal control tasks for system with integrators (Inverted Pendulum, Fig.~\ref{fig:exp1_samples_cost}(b)), AVWBFO (proposed) slightly outperforms MPPI, while WBFO (proposed) shows inferior performance, confirming the importance of discounted reward for dynamic systems.

\subsubsection{Franka Arm Collision Avoidance Planning}
\begin{figure}
    \centering
    \includegraphics[width=1\linewidth]{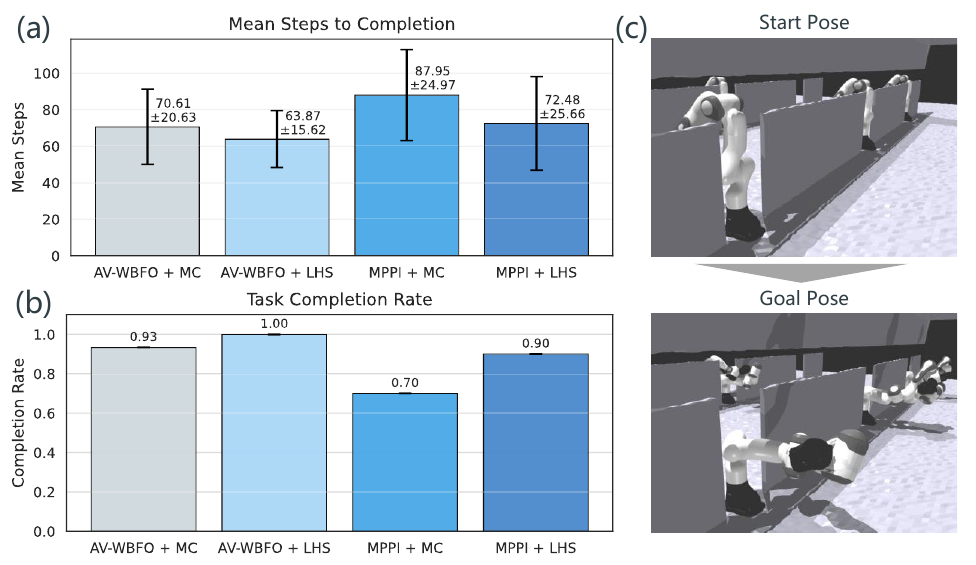}
    \caption{Franka planning task. 30 manipulators are planned using 4 different method combinations to reach a goal behind a wall gap. (a) Mean steps to completion and standard deviation over 30 trials. (b) Completion rates of each method within 150 steps. (c) Start and final states of the planning task.}
    \label{fig:franka_plan}
\end{figure}

We evaluate completion rates and efficiency of Franka arm collision avoidance planning across four method combinations: AVWBFO (proposed)/MPPI with Monte Carlo (MC) or Latin Hypercube Sampling (LHS). 30 Franka manipulators are tasked with reaching a goal behind a wall gap within 150 steps using 64 samples for all methods, as depicted in Fig.~\ref{fig:franka_plan}.

\textbf{Results.} AVWBFO (proposed) achieves higher completion rates than MPPI (93.3\% vs 70\% with MC sampling, 100\% vs 90\% with LHS). LHS sampling consistently improves performance, with AVWBFO (proposed)+LHS achieving optimal results (100\% completion rate, 63.9±15.6 steps to completion). This demonstrates that both optimization algorithm choice and sampling strategy contribute to improved planning performance.

\subsection{Robot Navigation Performance in Rough Terrain}

\begin{figure*}[t!]
    \centering
    \includegraphics[width=1\textwidth]{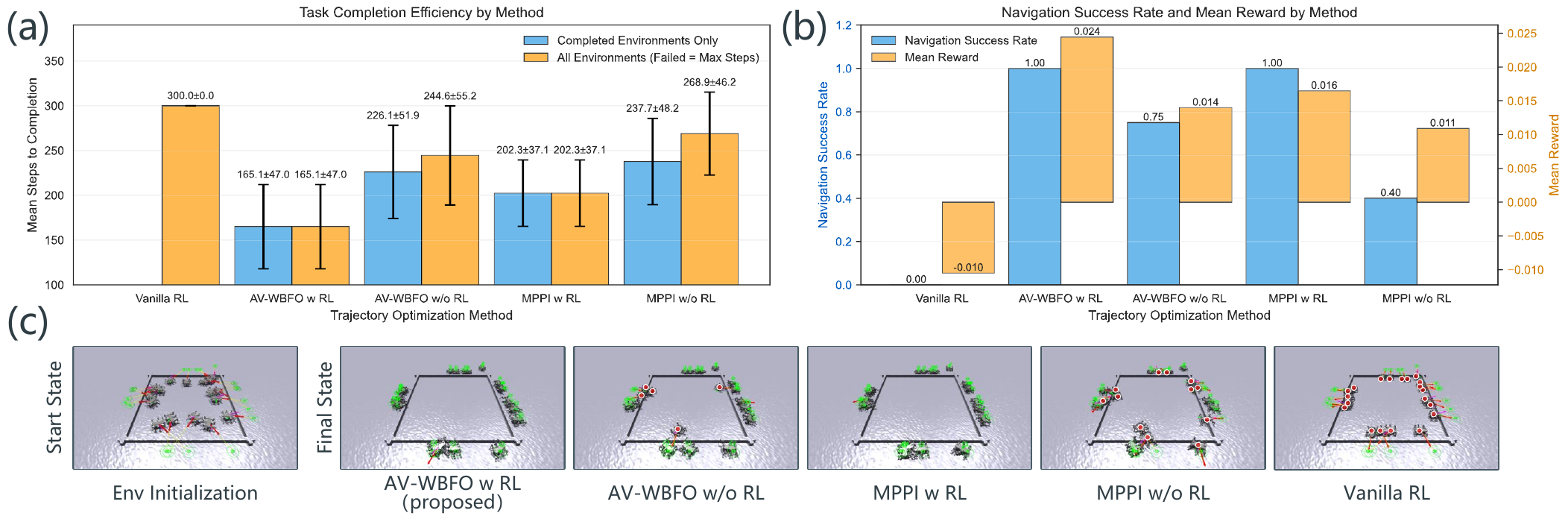}
    \caption{Hexapod Barrier Navigation. (a) Navigation task completion efficiency comparison. Mean steps to completion and standard deviation are measured over 20 trials. (b) Navigation success rate and mean reward during trajectory optimization. (c) The navigation task setup with 20 Hexapod robots starting inside a square barrier (height 0.2m, width 0.35m) and navigating to an external goal. The snapshots at 300 steps of each method are shown, with successful robots marked in green and failed ones in red.}
    \label{fig:exp2_elair_nav}
\end{figure*}

We validate the complete hierarchical rolling-denoising framework in realistic legged robot navigation scenarios with barrier crossing challenges. The experimental setup uses 20 Hexapod robots navigating from inside a square barrier (height 0.2m, width 0.35m) to an external goal within 300 steps (6 seconds at 0.02s step time) with 128 samples, testing agile maneuvering in constrained spaces, as shown in Fig.~\ref{fig:exp2_elair_nav}(c).

Five trajectory optimization approaches are compared: (1) \textbf{Vanilla RL} (pre-trained policy from flat terrain without trajectory optimization), (2) \textbf{AVWBFO w RL (proposed)} (our algorithm warm-started with RL policy), (3) \textbf{AVWBFO w/o RL} (our algorithm without RL warm-start), (4) \textbf{MPPI w RL} (MPPI with RL warm-start), and (5) \textbf{MPPI w/o RL} (MPPI without RL warm-start). Performance is evaluated using navigation success rate, mean completion steps, and accumulated reward.

\textbf{Results and Analysis.} The experimental results, as illustrated in Fig.~\ref{fig:exp2_elair_nav}, reveal three critical insights:

\begin{itemize}
    \item \textbf{RL Warm-starting Impact}: Both AVWBFO (proposed) w RL and MPPI w RL achieved perfect 100\% success rates (Fig.~\ref{fig:exp2_elair_nav}(b)), while their non-warm-started counterparts achieved only 75\% and 40\% respectively. Vanilla RL completely failed (0\% success), highlighting the necessity of trajectory optimization for complex navigation.

    \item \textbf{Algorithm Superiority}: AVWBFO (proposed) consistently outperformed MPPI in both configurations, as shown in Fig.~\ref{fig:exp2_elair_nav}(a). With RL warm-starting, AVWBFO (proposed) required 165.1±47.0 steps versus MPPI's 202.4±37.1 steps. Without warm-starting, AVWBFO (proposed) needed 226.1±51.9 steps versus MPPI's 237.7±48.2 steps.

    \item \textbf{Solution Quality}: Mean reward analysis (Fig.~\ref{fig:exp2_elair_nav}(b)) confirms trajectory optimality, with AVWBFO (proposed) w RL achieving the highest reward (0.024), followed by MPPI w RL (0.016), AVWBFO (proposed) w/o RL (0.014), MPPI w/o RL (0.011), and Vanilla RL (-0.010). Higher rewards indicate smoother, collision-free trajectories.
\end{itemize}

These results validate our hierarchical approach, demonstrating that the combination of learned behaviors with adaptive trajectory optimization significantly enhances both success rates and motion quality in challenging navigation scenarios.







\section{Conclusion}

In this paper, we introduced PegasusFlow, a hierarchical rolling-denoising framework that fundamentally addresses the expert data dependency bottleneck in diffusion-based robot planners. By enabling direct and parallel sampling of trajectory score gradients from environmental interaction, our approach reduces reliance on imitation learning, paving the way for training diffusion policies via pure score-matching.

Our core contribution, the WBFO algorithm and its action-value variant (AVWBFO), was shown to be significantly more sample-efficient and effective than traditional sampling-based methods like MPPI. Experiments demonstrated that combining AVWBFO with an RL warm-start yields a powerful planner capable of solving complex locomotion tasks. In a challenging barrier-crossing scenario (Fig.~\ref{fig:exp2_elair_nav}), our method achieved a 100\% success rate and was 18\% faster than the next-best baseline, highlighting its suitability for agile navigation in constrained environments where pure RL policies fail.

By bridging the gap between the generative power of diffusion models and the practical demands of real-time robotic control, this work opens new avenues for developing highly capable and autonomous legged robots. Future work will focus on deploying this framework for full diffusion policy training and extending it to multi-robot coordination and dynamic environments.

\bibliographystyle{IEEEtran}
\bibliography{library}

\end{document}